\documentclass[letterpaper]{article} 
\usepackage{aaai25}  
\usepackage{times}  
\usepackage{helvet}  
\usepackage{courier}  
\usepackage[hyphens]{url}  
\usepackage{graphicx} 
\usepackage{natbib}  
\usepackage{caption} 
\usepackage{algorithm}
\usepackage{algorithmic}
\usepackage{booktabs}
\usepackage{multirow}

\usepackage{newfloat}
\usepackage{listings}
\DeclareCaptionStyle{ruled}{labelfont=normalfont,labelsep=colon,strut=off} 
\floatstyle{ruled}
\newfloat{listing}{tb}{lst}{}
\floatname{listing}{Listing}
\title{Transformer-Based Approach for \\Automated Functional Group Replacement in Chemical Compounds}
\author{
    Bo Pan\textsuperscript{\rm 1},
    Zhiping Zhang\textsuperscript{\rm 2},
    Kevin Spiekermann\textsuperscript{\rm 2},
    Tianchi Chen\textsuperscript{\rm 2}, 
    Xiang Yu\textsuperscript{\rm 2}, \\
    Liying Zhang\textsuperscript{\rm 2}$^{*}$,
    Liang Zhao\textsuperscript{\rm 1}$^{*}$
}
\affiliations{
    \textsuperscript{\rm 1}Emory University, Atlanta, GA, USA\\
    \textsuperscript{\rm 2}Merck \& Co., Inc., Rahway, NJ, USA\\
    \texttt{liying.zhang@merck.com, liang.zhao@emory.edu}
}

\usepackage{bibentry}

\begin{document}
\copyrighttext{}

\maketitle
\renewcommand{\thefootnote}{\fnsymbol{footnote}}  
\footnotetext[1]{Corresponding authors.}

\begin{abstract}
Functional group replacement is a pivotal approach in cheminformatics to enable the design of novel chemical compounds with tailored properties. Traditional methods for functional group removal and replacement often rely on rule-based heuristics, which can be limited in their ability to generate diverse and novel chemical structures. Recently, transformer-based models have shown promise in improving the accuracy and efficiency of molecular transformations, but existing approaches typically focus on single-step modeling, lacking the guarantee of structural similarity. In this work, we seek to advance the state of the art by developing a novel two-stage transformer model for functional group removal and replacement. Unlike one-shot approaches that generate entire molecules in a single pass, our method generates the functional group to be removed and appended sequentially, ensuring strict substructure-level modifications. Using a matched molecular pairs (MMPs) dataset derived from ChEMBL, we trained an encoder-decoder transformer model with SMIRKS-based representations to capture transformation rules effectively.
Extensive evaluations demonstrate our method’s ability to generate chemically valid transformations, explore diverse chemical spaces, and maintain scalability across varying search sizes. 

\end{abstract}

%

\section{Introduction}

Chemical compound design is an essential task in drug discovery, where the ability to modify molecular structures plays a critical role in improving the properties of substances. One of the most challenging tasks in this domain is proposing new drug candidates while maintaining the overall structure and properties of the compound. This can be achieved by the replacement of functional groups in molecules. Functional groups—specific atoms or clusters of atoms that impart distinct chemical properties to a molecule—are often targeted for modification to tune a compound's biological, chemical, or physical properties. In drug discovery, for instance, altering functional groups can lead to more potent or selective drugs, whereas in materials science, modifying functional groups can affect properties like solubility, stability, and conductivity.

Despite its importance, functional group replacement is traditionally a labor-intensive process, often relying on expert chemists who manually select replacement groups based on their experience and chemical intuition. While rule-based methods exist, they are limited by their inability to scale and explore large, complex chemical spaces. The development of more scalable, data-driven methods has been a focus in recent years, and machine learning models—particularly deep learning techniques such as transformers—have shown remarkable potential in learning complex molecular relationships from large datasets. These machine learning-based approaches can be broadly categorized into two streams: (1) {one-shot methods} \cite{jin2018learning, tysinger2023can, tibo2024exhaustive, he2022transformer, he2021molecular} and (2) {two-stage methods} \cite{wu2024leveraging, chen2021molecule, jin2020multi, wang2024efficient}. 
{One-shot methods} aim to generate an entirely new compound in a single pass, given an input compound. 
Early work in this space largely relied on VAE-based architectures to perform molecule-level translation, and the constraint of functional-group change is enforced by choosing input–output molecule pairs based on structural similarity\cite{jin2018learning}. 
Although VAE-based models ensured a baseline level of similarity through their data construction, their performance was hindered by limited scalability when the data volume gets large. 
More recently, transformer-based models have been employed to address these limitations, leveraging their capacity to handle large training datasets effectively \cite{tysinger2023can, tibo2024exhaustive, he2022transformer, he2021molecular}. 
However, despite these advancements, one-shot methods continue to struggle with strictly enforcing similarity between the generated and source molecules. This limitation arises because these models rely solely on training data to enforce functional group changes, lacking a model-level guarantee for such constraints.
In contrast, two-stage methods decompose the problem into two sequential steps: first identifying the functional group to be removed and then proposing a new functional group to append at the attachment point. 
Early implementations of two-stage methods often utilized graph-based models for these sub-tasks \cite{chen2021molecule, jin2020multi}. 
Recent advances \cite{wu2024leveraging, wang2024efficient} have incorporated large-scale pretrained language models (LLMs) to suggest functional group replacements directly through In-Context Learning (ICL)\cite{brown2020language}.
However, ICL-based methods are constrained by limited examples in the prompt, restricting their ability to process and generate replacements based on a wide range of examples. 

In this paper, we present a novel approach to functional group replacement that leverages transformer models and adopts a two-stage generation strategy. Our method involves two sequential tasks: first, predicting which functional group to remove from the source molecule, and second, selecting a new functional group to append at the specified location. This approach, which is grounded in a data-driven, autoregressive framework, enables a scalable and robust solution for functional group replacement. By training on large chemical datasets, our method ensures the generation of diverse, valid, and chemically relevant compounds. Furthermore, our model supports both user-specified transformations as well as those suggested by the model itself, offering flexibility for a wide range of applications. We demonstrate the performance and efficacy of our approach through a series of experiments, where we show that it can generate compounds that are both novel and chemically plausible, contributing to a deeper understanding of functional group modification.

\section{Related Work}
\subsection{Matched Molecular Pairs}
Matched molecular pair (MMP) analysis is a cheminformatic methodology focusing on the systematic exploration of chemical structure-property relationships (SAR). MMPs are defined as pairs of compounds that differ by a localized structural change, enabling targeted insights into how specific modifications influence molecular properties. By abstracting transformations into structured rules, MMP analysis facilitates the study of structure-activity relationships, supporting diverse applications such as bioisostere identification, optimization of physicochemical properties, and prediction of biological activity. Modern computational platforms, such as mmpdb, leverage advanced algorithms for fragment-and-index processing, allowing researchers to identify, catalog, and analyze MMPs from extensive datasets with high efficiency \cite{dalke2018mmpdb, hussain2010computationally}. The increasing availability of large chemical datasets, coupled with advancements in data mining techniques, has further cemented the role of MMP analysis in drug discovery and molecular design. 

\subsection{Machine Learning-based Functional Group Replacement}
Current machine learning-driven approaches can be broadly classified into two categories: (1) one-shot methods \cite{jin2018learning, tysinger2023can, tibo2024exhaustive, he2022transformer, he2021molecular} and (2) two-stage methods \cite{wu2024leveraging, chen2021molecule, jin2020multi, wang2024efficient}. One-shot methods aim to generate a completely new compound from an input compound in a single step. Early research in this area primarily relied on Variational Autoencoder (VAE)-based architectures, which facilitated molecule-level translation by enforcing functional group changes through the selection of input-output molecule pairs based on structural similarity \cite{jin2018learning}. Recent advancements have introduced transformer-based models to handle large training datasets \cite{tysinger2023can, tibo2024exhaustive, he2022transformer, he2021molecular}. Two-stage methods approach the problem in two distinct steps: first, identifying the functional group to be removed, followed by proposing a new functional group for attachment. Existing study include graph-based models \cite{chen2021molecule, jin2020multi}. and LLM-based methods \cite{wu2024leveraging, wang2024efficient, brown2020language}.

\section{Methods}

In this work, we introduce a transformer-based two-stage generation approach for functional group replacement in chemical compounds. Our method builds on the autoregressive generation capabilities of transformer models, allowing us to model the sequential nature of molecular transformations while ensuring precise control over the functional groups being replaced.

The core idea of our approach involves generating a transformation string in the SMIRKS format, which encodes both the removal of the original functional group and the addition of the new functional group. We first leverage the transformer model to identify the functional group to be removed from the input molecule. Once this functional group is identified, the model proceeds to generate the corresponding functional group to append. This structured approach ensures that only the functional group level is modified, and the overall molecular structure is maintained, which is crucial for ensuring that the generated compounds are valid and chemically meaningful.

\begin{figure*}[ht]
    \centering
    \includegraphics[width=0.65\textwidth]{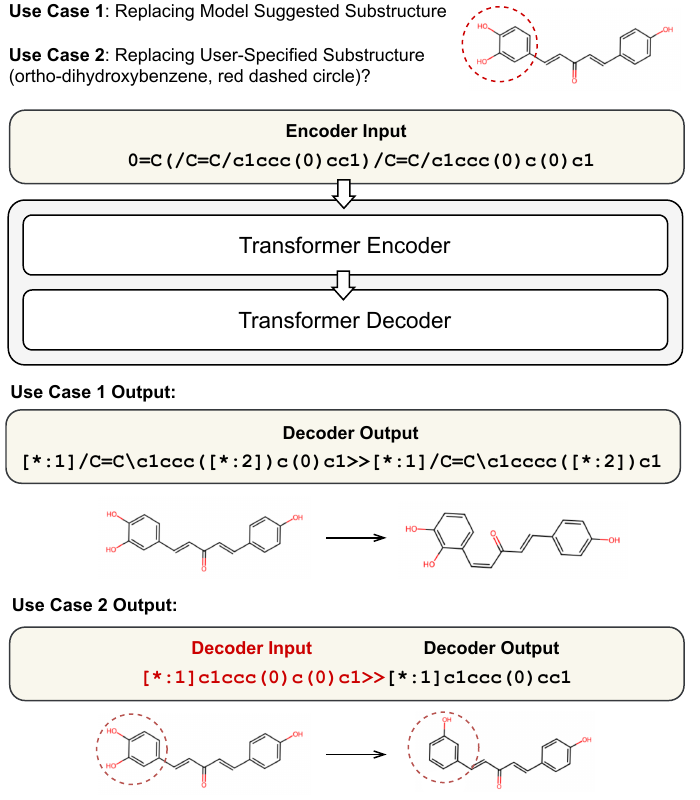}
    \caption{An illustration of our method, including two use cases: 1) replacing model-suggested substructure and 2) replacing user-specified substructure.}
    \label{fig:method}
\end{figure*}

As illustration of our method is given in Fig.~\ref{fig:method}, our method can handle two types of replacement tasks: (1) model-suggested replacements, where the model identifies optimal functional group changes based on the learned patterns from the data, and (2) user-specified replacements, where the user defines the functional group to be replaced, and the model generates the corresponding transformation. By incorporating both of these tasks, our approach offers flexibility for a wide range of use cases in drug discovery, materials design, and other chemical engineering fields.

\subsection{Data Collection}

We collected data from the ChEMBL database \cite{gaulton2012chembl}, a comprehensive resource of bioactive molecules with drug-like properties. ChEMBL currently contains over 2 million distinct compounds. To extract matched molecular pairs (MMPs), we utilized MMPDB \cite{dalke2018mmpdb}, a tool that identifies molecular pairs from a database based on structural similarities. MMPDB works by detecting a core structure shared between molecules and defining transformations at the R-groups. The parameters were configured as follows: the number of heavy atoms in the core was set to $\leq$ 50, the number of heavy atoms in the R-group to $\leq$ 13, and the ratio of heavy atoms in the R-group to the total molecule to $<$ 0.33. By running MMPDB on ChEMBL, we got 465 million MMPs. We further filtered the data by constraining that each molecule has at most 10 transformations to prevent highly popular molecules from dominating the training dataset. Additionally, each transformation string was limited to appear no more than 10 times. Eventually, we randomly sampled 2 million MMPs from the data filtered with the above processes as our training data. 

For our model, we used the source molecule as the input and the corresponding transformation string as the label. An example of input-output pair of our training data is as follows: for a molecule
$$
\verb|0=C(C=Cc1ccc(0)cc1)C=Cc1ccc(0)c(0)c1|
$$

a possible transformation string could be
$$
\verb|[*:1]c1ccc(0)c(0)c1>>[*:1]c1ccc(0)cc1| 
$$
(dehydroxylation transformation in SMIRKS format). 

\subsection{Modeling}
We employ the encoder-decoder transformer architecture \cite{raffel2020exploring} as our base model architecture. Encoder-decoder models are particularly advantageous for this task as they allow explicit modeling of input-output pairs, enabling the model to effectively capture the complex relationships between molecular structures and their transformation processes. Unlike decoder-only models optimized for autoregressive tasks, encoder-decoder models process the entire input sequence to generate a detailed representation, enabling accurate molecular transformation learning. This allows encoder-decoder architectures to be better for fine-tuning on specific tasks \cite{fu2023decoder}.

\subsection{Inference}
As illustrated in Fig.~\ref{fig:method}, after training, the model supports two approaches for performing inference: 1) \textbf{Replacing user-specified substructures}, where the user defines the target substructure to be replaced. This involves providing the model with a molecule to be modified directly as input, allowing it to predict transformations autonomously. 2) \textbf{Replacing model-suggested substructures}, where the model identifies optimal replacements based on its learned patterns. This involves passing the model with a molecule as input to the encoder and a functional group representation (e.g., \texttt{[O:1]}) as input to the decoder. This decoder input forces the model to generate a SMIRKS transformation string that begins with the specified pattern, ensuring the user-defined functional group is replaced in the generated transformation.
Our method leverages the transformer's probabilistic nature during inference to generate multiple plausible replacements using beam search.

\begin{table*}[t]
\centering
\begin{tabular}{@{}clccccc@{}}
\toprule
\multirow{2}{*}{Model} & \multirow{2}{*}{Metric} & \multicolumn{5}{c}{Search size (k)} \\ \cmidrule(l){3-7} 
 &  & 1 & 10 & 20 & 50 & 100 \\ \midrule

\multirow{2}{*}{Mol2Mol model} 
 & \%Valid & 1.0000 & 1.0000 & 1.0000 & 0.9984 & 0.9497 \\
 & \%Exist & \textbf{0.9949} & 0.8795 & 0.8272 & 0.7235 & 0.6244 \\ \midrule

\multirow{2}{*}{Mol2Trans model} 
 & \%Valid & 1.0000 & 0.9974 & 0.9903 & 0.9713 & 0.9391 \\
 & \%Exist & 0.9692 & \textbf{0.9399} & \textbf{0.9045} & \textbf{0.8315} & \textbf{0.7377} \\ 

\bottomrule
\end{tabular}
\caption{Performance comparison of Mol2Mol and Mol2Trans models.}
\label{tab:model_performance}
\end{table*}

\section{Experiments}
For our experiments, we focus on evaluating the performance of our transformer-based two-stage functional group replacement model against traditional methods and existing state-of-the-art models. We set up the following three key tasks:

Functional Group Removal: The first task involves the model's ability to correctly identify and predict which functional group is to be removed from the input molecule. We define the target functional group by selecting a specific functional group from the input molecule and removing it. The model is then tasked with predicting this removal accurately.
Functional Group Replacement: In the second task, once the functional group is removed, the model predicts an appropriate replacement functional group from a predefined set of candidates. These candidates include commonly used groups such as hydroxyl, amine, carbonyl, and halogens, among others. The replacement functional group must be added to the molecule at the same location from which the previous group was removed, ensuring that the output remains chemically valid.
Overall Transformation: The final task evaluates the model’s ability to perform both removal and replacement in a single, end-to-end transformation. This task tests the model's overall ability to generate valid molecules with the specified modifications.
Each task is evaluated based on several key metrics, including chemical validity, transformation accuracy, and molecular novelty. Chemical validity is assessed using a cheminformatics tool that checks for syntactical correctness in SMILES representation. Transformation accuracy measures how well the model performs the intended removal and replacement of functional groups. Molecular novelty is calculated by comparing the generated molecules to the training set to ensure that the model is not simply memorizing the data.

\subsection{Implementation Details}

\subsubsection{Model Architecture}

To leverage a model backbone with a pre-existing understanding of molecular languages, we leverage the ChemT5 model \cite{christofidellis2023unifying}, an encoder-decoder transformer pretrained on chemical datasets, as our base model. ChemT5 contains approximately 220 million parameters and has been fine-tuned for tasks in cheminformatics. It consists of 12 layers, 12 attention heads, 220M parameters, and processes input sequences up to 512 tokens \cite{christofidellis2023unifying}.
\subsubsection{Training and Inference}
We employed standard supervised training to fine-tune all parameters of the base model. Teacher forcing \cite{williams1989learning} is incorporated to improve training stability. The training was conducted with a batch size of 64 on each device, and a learning rate of 5e-4. We use an early stop strategy with a tolerance of 2 epochs based on the evaluation loss. Utilizing four NVIDIA A6000 GPUs (48 GB each), the training process required approximately 70 hours to complete. During inference, we use a beam search size of 100 and a temperature of 0.3.

\subsection{Coverage of Ground Truth Replacements} 

We evaluated our proposed method on the held-out test set by generating top-\emph{k} predictions (where \emph{k} = 1, 10, 20, 50, and 100) for each input molecule, using a beam search size of 200. Two metrics were measured for each set of predictions: (1) the percentage of generated SMIRKS that can be successfully applied to the input molecule to yield a valid product (\%Valid), and (2) the percentage of generated products that already exist in our dataset (\%exist), indicating that they remain within a drug-like chemical space.

Table \ref{tab:model_performance} compares the performance of two models trained on the same dataset but with different output formats—\textbf{Mol2Mol} (directly translating a source compound into a target compound) versus our proposed approach \textbf{Mol2Trans}. Overall, both models achieve high \%Valid scores across all \emph{k} values, showing their ability to generate chemically valid transformations. For instance, at \emph{k}=1, both methods produce essentially 100\% valid outputs, confirming that their top-ranked predictions rarely fail chemically.

The two models differ, however, in how many of their generated molecules overlap with the training set. At \emph{k}=1, Mol2Mol slightly outperforms Mol2Trans in \%exist (99.49\% vs. 96.92\%), suggesting that its top prediction is more likely to reproduce known molecules. As \emph{k} increases, the gap in \%Valid between the two approaches remains small (e.g., at \emph{k}=100, Mol2Mol reaches 94.97\% while Mol2Trans achieves 93.91\%). Yet for \%exist at \emph{k}=100, Mol2Trans (73.77\%) surpasses Mol2Mol (62.44\%), implying that Mol2Trans is better at sampling known, “drug-like” molecules among its broader predictions.

These results demonstrate that (1) both models can reliably generate valid chemical transformations, even when exploring a large beam search size of 200; and (2) Mol2Trans maintains a strong balance between valid transformation and producing compounds that remain within familiar chemical space as \emph{k} grows. In scenarios emphasizing near-perfect retention within known scaffolds, Mol2Mol provides slightly higher fidelity at very low \emph{k}, whereas Mol2Trans offers a more consistent \%exist across a broader set of predictions when a larger number of candidate outputs is desired.

\subsection{Performance on Different Frequency Levels of Source Molecules}

We split our test set into multiple groups based on the number of matched molecules (\emph{k}) each source molecule has in the dataset. For instance, the group with $k=10$ consists of source molecules that each have exactly ten known matched transformations in the dataset. This grouping reflects how “popular” or “frequently available” each source molecule is within the collection. To evaluate coverage, we generate candidate products for each source molecule in a given group, then measure: \textbf{Coverage Rate}: The percentage of source molecules for which at least one of the known matched products was reproduced among our generated results. \textbf{Avg.\ \# Covered}: The average number of known matched products (out of $k$) that appear in the generation set for each source.

Table~\ref{tab:mmp_coverage} summarizes the coverage metrics for each group. With $k=1$, our method covers 73.33\% of known transformations; intuitively, there is only one possible target molecule to retrieve, so an average of 0.73 out of 1 is covered. As $k$ increases, we observe different trends in coverage. Notably, for $k=10$, the coverage rate peaks at 79.00\%, with an average of 7.9 out of 10 possible targets found in the generated sets. For $k=20$, the coverage rate is 71.33\%, indicating that while there are more potential matches, the method still captures an average of 14.2. Interestingly, coverage does not consistently decrease or increase with higher $k$; for instance, $k=50$ yields 68.08\% coverage, which is higher than $k=40$. These results suggest that our approach remains robust even when the number of possible matched transformations grows, successfully recovering a substantial fraction of known products for each source molecule.  

\begin{table}[t]
\centering
\begin{tabular}{@{}ccc@{}}
\toprule
\textbf{\# Target Mols} & \textbf{Coverage Rate} & \textbf{Avg.\ \# Covered} \\ 
\midrule
1   & 73.33\% & 0.73 \\
10  & 79.00\% & 7.9 \\
20  & 71.33\% & 14.2 \\
30  & 63.33\% & 19.0 \\
40  & 60.33\% & 24.1 \\
50  & 68.08\% & 34.0 \\
\bottomrule
\end{tabular}
\caption{\textbf{Coverage of MMPs in the dataset for different values of \emph{k}.} Each group consists of 100 examples, where $k$ is the number of matched molecules known in the dataset for each source. “Avg.\ \# Covered” indicates the average subset of the $k$ matched molecules that appears in our generated results.}
\label{tab:mmp_coverage}
\end{table}

\subsection{Performance under Different Search Size}
\begin{figure}[t]
    \centering
    \includegraphics[width=0.40\textwidth]{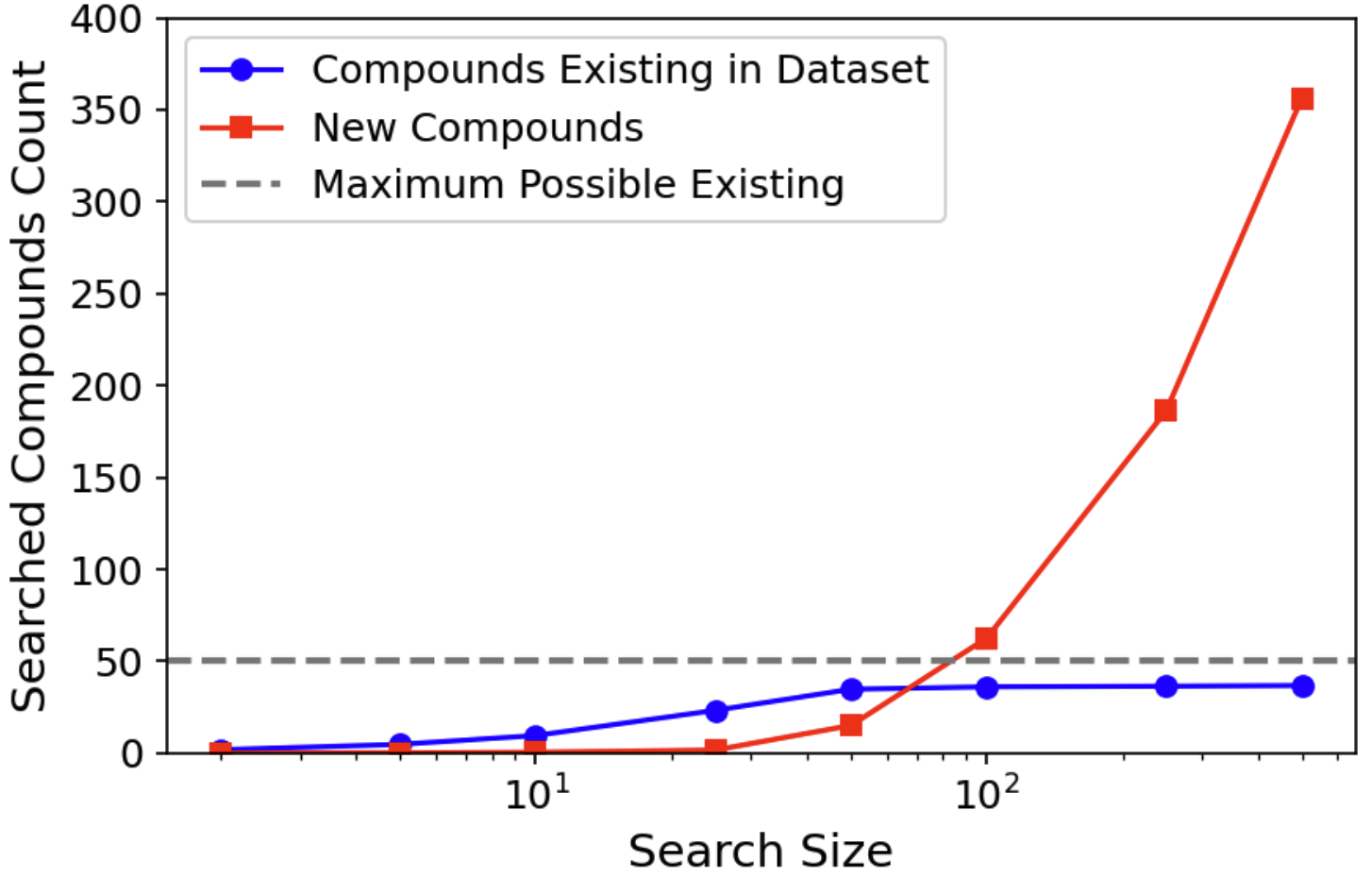} 
    \caption{The searched compounds that are existing in the dataset (red) and unseen in the dataset (blue) under different search sizes.}
    \label{fig:curve}
\end{figure}

We further evaluate our model by varying the beam search size ($k$) to investigate how it balances reproducing known molecules versus exploring new chemical structures. Specifically, we extract a subset of test examples where each source molecule has exactly 50 known target molecules within our dataset. This setup allows us to measure: \textbf{Existing (Known) Compounds in the dataset}: Among the $k$ generated candidates, the number of generated that match these 50 targets. \textbf{New (Novel) Compounds}: Among the $k$ candidates, the number that are not present in our dataset (thus expanding the chemical space).  We experiment with a range of $k$ values (e.g., $k \in \{1, 5, 10, 50, 100, 200, 500, 1000\}$), generating that many candidate outputs for each source molecule. By examining how the balance of existing versus novel compounds changes with $k$, we gain insight into whether our model is merely memorizing learned transformations or truly exploring new regions of chemical space.

Figure~\ref{fig:curve} illustrates our findings. As $k$ increases, the total number of generated products grows, naturally allowing the model to retrieve more of the 50 known targets. For small $k$ (e.g., 1 or 5), only a few of these known molecules are recovered, indicating that the model focuses on high-confidence predictions. Once $k$ surpasses 10, however, we observe a substantial rise in recovered known compounds (blue dots), highlighting that the learned representation is effective at recalling multiple valid transformations from the dataset. Concurrently, we also see an increasing number of novel compounds (red dots) as $k$ grows. This suggests that the model goes beyond the strict confines of the training data, proposing new chemical entities that may be interesting starting points for further exploration. Notably, the rate at which novel compounds appear accelerates significantly as $k$ becomes large, suggesting that the model’s generative capacity extends well beyond memorized patterns. Overall, these results demonstrate that our approach can be tuned via the search size $k$ to emphasize either (1) higher precision on known transformations for smaller $k$, or (2) broader exploration of chemical space with larger $k$. Striking a balance between these two extremes is beneficial for practical applications, as it ensures both confidence in generating known drug-like motifs and the capacity to discover new, potentially innovative molecules.

\section{Discussion}

\textbf{Conclusion.} Our current method is the first large-scale machine learning model to ensure functional group replacement by a two-stage generation paradigm. Our method can suppose diverse functional group replacements for the input source molecule, supporting both replacing model-suggested functional groups and user-specified ones. Our model can generate a large percentage (approximately 70\%) of the compounds covered in the dataset, as well as suggesting diverse novel replacements unseen in the dataset by enlarging the search sizes, while ensuring 100\% of the generated compounds are only with a functional group-level change to the source compound.

\textbf{Limitations.} Despite promising, our current method falls short in the training data scale, compared to other state-of-the-art methods. For example, \cite{tibo2024exhaustive} leverages 200 billion compound pairs to train the transformer model, resulting in a model with better capabilities in searching more valid compounds in the local chemical space. Another drawback of our current method is that it is a property-agnostic method, which does not support property optimization guided generation. This may limit its application in real-world scenarios.

\textbf{Future directions.} Some future directions include: 1) Larger-scale training. The maximum possible data pairs we can attain from ChEMBL is 200 million pairs, while we are currently using 2 million. Training with more resources can be expected to continuously increase the model performance. 2) Property-conditioned generation. Starting from the current model, property-optimization generation can be implemented via either reinforcement learning-based fine-tuning or conditional generation-based fine-tuning. Appending property control to the model can be expected to make it more usable.
3) Different molecular representation methods. SELFIES (Self-Referencing Embedded Strings) \cite{krenn2020self} is a molecular representation where every generated string corresponds to a valid structure. This ensures validity in \textit{de novo} molecular design without needing post-generation checks, simplifying chemical space exploration and enhancing the efficiency of molecular generation algorithms.


\bibliography{molgen}

\end{document}